\documentstyle[aaai1]{article}

 \twocolumn 

\title{Machine Learning of Generic and User-Focused Summarization}

\author{Inderjeet Mani and Eric Bloedorn \\ The MITRE
Corporation, \\ 11493 Sunset Hills Road, W640, Reston, VA 22090, USA \\
\{imani,bloedorn\}@mitre.org}

\begin{document}

\bibliographystyle{named}

\maketitle

\begin{abstract}
\begin{quote}

A key problem in text summarization is finding a salience
function which determines what information in the source should be
included in the summary. This paper describes the use of machine
learning on a training corpus of documents and their abstracts to
discover salience functions which describe what combination of
features is optimal for a given summarization task. The method addresses
both ``generic'' and user-focused summaries.
\end{quote}
\end{abstract}

\section{Introduction\footnotemark}

\footnotetext{Copyright \copyright 1998, American Association for
Artificial Intelligence ({\tt www.aaai.org}). All rights
reserved.}

With the mushrooming of the quantity of on-line text information,
triggered in part by the growth of the World Wide Web, it is
especially useful to have tools which can help users digest
information content.  Text summarization attempts to address this need
by taking a partially-structured source text, extracting information
content from it, and presenting the most important content to the user
in a manner sensitive to the user's or application's needs.  The end
result is a condensed version of the original. A key problem in
summarization is determining what information in the source should be
included in the summary. This determination of the salience of
information in the source (i.e., a salience function for the text)
depends on a number of interacting factors, including the nature and
genre of the source text, the desired compression (summary length as a
percentage of source length), and the application's information
needs. These information needs include the reader's interests and
expertise (suggesting a distinction between ``user-focused'' versus
``generic'' summaries), and the use to which the summary is being put,
for example, whether it is intended to alert the user as to the source
content (the ``indicative'' function), or to stand in place of the
source (the ``informative'' function), or even to offer a critique of
the source (the ``evaluative'' function
\cite{Sparck-Jones}).

A considerable body of research over the last forty years has explored
different levels of analysis of text to help determine what
information in the text is salient for a given summarization task. The
salience functions are usually sentence filters, i.e.  methods for
scoring sentences in the source text based on the contribution of
different features.  These features have included, for example,
location \cite{Edmundson}, \cite{Paice}, statistical measures of term prominence
\cite{Luhn}, \cite{Brandow}, rhetorical structure \cite{Miike},
\cite{Marcu}, similarity between sentences \cite{Skorokhodko}, 
presence or absence of certain syntactic features
\cite{Pollock}, presence of proper names \cite{Kupiec}, and measures of
prominence of certain semantic concepts and relationships
\cite{Paice}, \cite{Maybury}, \cite{Fum}. In general, it appears that
a number of features drawn from different levels of analysis may
combine together to contribute to salience. Further, the importance of
a particular feature can of course vary with the genre of text.

Consider, for example, the feature of text location.  In newswire
texts, the most common narrative style involves lead-in text which
offers a summary of the main news item.  As a result, for most
varieties of newswire, summarization methods which use leading text
alone tend to outperform other methods \cite{Brandow}. However, even
within these varieties of newswire, more anecdotal lead-ins, or
multi-topic articles, do not fare well with a leading text approach
\cite{Brandow}. In other genres, other locations are salient: for
scientific and technical articles, both introduction and conclusion
sections might contain pre-summarized material; in TV news broadcasts,
one finds segments which contain trailing information summarizing a
forthcoming segment. Obviously, if we wish to develop a summarization
system that could adapt to different genres, it is important to have
an automatic way of finding out what location values are useful for
that genre, and how it should be combined with other features. Instead
of selecting and combining these features in an adhoc manner, which
would require re-adjustment for each new genre of text, a natural
suggestion would be to use machine learning on a training corpus of
documents and their abstracts to discover salience functions which
describe what combination of features is optimal for a given
summarization task. This is the basis for the trainable approach to
summarization.

Now, if the training corpus contains ``generic'' abstracts (i.e.,
abstracts written by their authors or by professional abstractors with
the goal of dissemination to a particular - usually broad - readership
community), the salience function discovered would be one which
describes a feature combination for generic summaries. Likewise, if
the training corpus contains ``user-focused'' abstracts, i.e.,
abstracts relating information in the document to a particular user
interest, which could change over time, then then we learn a function
for user-focused summaries. While ``generic'' abstracts have
traditionally served as surrogates for full-text, as our computing
environments continue to accommodate increased full-text searching,
browsing, and personalized information filtering, user-focused
abstracts have assumed increased importance.  Thus, algorithms which
can learn both kinds of summaries are highly relevant to current
information needs. Of course, it would be of interest to find out what
sort of overlap exists between the features learnt in the two
cases. 

In this paper we describe a machine learning approach which learns
both generic summaries and user-focused ones. Our focus is on machine
learning aspects, in particular, performance-level comparison between
different learning methods, stability of the learning under different
compression rates, and relationships between rules learnt in the
generic and the user-focused case.

\section {Overall Approach}

In our approach, a summary is treated as a representation of the
user's information need, in other words, as a query. The training
procedure assumes we are provided with training data consisting of a
collection of texts and their abstracts. The training procedure first
assigns each source sentence a relevance score indicating how relevant
it is to the query. In the basic ``boolean-labeling'' form of this
procedure, all source sentences above a particular relevance threshold
are treated as ``summary'' sentences. The source sentences are
represented in terms of their feature descriptions, with ``summary''
sentences being labeled as positive examples. The training sentences
(positive and negative examples) are fed to machine learning
algorithms, which construct a rule or function which labels any new
sentence's feature vector as a summary vector or not. In the generic
summary case, the training abstracts are generic: in our corpus they
are author-written abstracts of the articles. In the user-focused
case, the training abstract for each document is generated
automatically from a specification of a user information need.

It is worth distinguishing this approach from other previous work in
trainable summarization, in particular, that of \cite{Kupiec} at
Xerox-Parc (referred to henceforth as Parc), an approach which has
since been followed by \cite{Teufel}. First, our goal is to learn
rules which can be easily edited by humans. Second, our approach is
aimed at both generic summaries as well as user-focused summaries,
thereby extending the generic summary orientation of the Parc work.
Third, by treating the abstract as a query, we match the entire
abstract to each sentence in the source, instead of matching
individual sentences in the abstract to one or more sentences in the
source.  This tactic seems sensible, since the distribution of the
ideas in the abstract across sentences of the abstract is not of
intrinsic interest. Further, it completely avoids the rather tricky
problem of sentence alignment (including consideration of cases where
more than one sentence in the source may match a sentence in the
abstract), which the Parc approach has to deal with. Also, we do not
make strong assumptions of independence of features, which the Parc
based work which uses Bayes' Rule does assume. Other trainable
approaches include \cite{Lin}; in that approach, what is learnt from
training is a series of sentence positions.  In our case, we learn
rules defined over a variety of features, allowing for a more abstract
characterization of summarization. Finally, the learning process does
not require any manual tagging of text; for ``generic'' summaries it
requires that ``generic'' abstracts be available, and for user-focused
abstracts, we require only that the user select documents that match
her interests.

\begin
{table*}[t]
\centering
{\small

\begin{tabular}{|l|l|l|}\hline
\multicolumn{3}{|c|}{\bf Location Features \/} \\ \hline
{\bf Feature \/} &   {\bf Values \/} &     {\bf Description \/} \\ \hline
{\bf sent-loc-para \/}
   &     \{1, 2, 3\}  & sentence occurs in first, middle  or
last third of para \\ \hline
{\bf  para-loc-section
\/}     &     \{1, 2, 3\}  & sentence occurs in first, middle
or last third of  section \\ \hline
{\bf
sent-special-section  \/}     &     \{1, 2, 3\}  & 1 if sentence occurs
in introduction, 2 if in 
conclusion, 3 if in other \\ \hline
 {\bf
depth-sent-section   \/}     &     \{1, 2, 3, 4\}  & 1 if sentence is a
top-level section, 4 if sentence is a subsubsubsection \\ \hline
\multicolumn{3}{|c|}{\bf Thematic Features \/} \\ \hline
{\bf
sent-in-highest-tf  \/}     &     \{1, 0\}  &  
average tf score (Filter 1) \\ \hline
{\bf
sent-in-highest-tf.idf  \/}     &     \{1, 0\}  &  
average tf.idf score (Filter 1)\\ \hline
{\bf
sent-in-highest-G$^{2}$  \/}     &     \{1, 0\}  &  
average $G^{2}$ score (Filter 1) \\  \hline
{\bf sent-in-highest-title  \/}     &     \{1, 0\}  &  
number of section heading or title term mentions (and
Filter 1) \\ \hline
{\bf
sent-in-highest-pname  \/}     &     \{1, 0\}  &  
number of  name mentions (Filter 1) \\ \hline
\multicolumn{3}{|c|}{\bf Cohesion Features \/} \\ \hline
{\bf sent-in-highest-syn \/} & \{1, 0\}  & number of unique sentences
 with a synonym link to sentence (Filter 1) \\ \hline
{\bf
sent-in-highest-co-occ  \/}     &     \{1, 0\} & number of unique
sentences
 with a  co-occurrence link to sentence (Filter 1) \\  \hline
\end{tabular}}
\caption{Text Features}
\label{table-1}
\end{table*}

\section{Features}

The set of features studied here are encoded as in Table~\ref{table-1}, where
they are grouped into three classes. {\it Location} features exploit the
structure of the text at different (shallow) levels of analysis. Consider the
{\it Thematic} features\footnote{Filter 1 sorts all the sentences in the
document by the feature in question. It assigns 1 to the current sentence iff
it belongs in top $c$ of the scored sentences, where $c$ = compression rate. As
it turns out, removing this discretization filter completely, to use raw scores
for each feature, merely increases the complexity of learnt rules without improving
performance}. The feature based on proper names is extracted using SRA's
NameTag \cite{Krupka}, a MUC6-fielded system. We also use a feature based on
the standard tf.idf metric : the weight $dw(i, k, l)$ of term $k$ in document
$i$ given corpus $l$ is given by: {\small \[ dw(i, k, l) =
tf_{ik}.(\ln(n)-\ln(df_{k}) +1) \]} where $tf_{ik}$ = frequency of term $k$ in
document $i$ divided by the maximum frequency of any term in document $i$,
$df_{k}$ = number of documents in $l$ in which term $k$ occurs, $n$ = total
number of documents in $l$. While the tf.idf metric is standard, there are some
statistics that are perhaps better suited for small data sets \cite{Dunning}.
The $G^{2}$ statistic indicates the likelihood that the frequency of a term in
a document is greater than what would be expected from its frequency in the
corpus, given the relative sizes of the document and the corpus. The version we
use here (based on \cite{Cohen}) uses the raw frequency of a term in a
document, its frequency in the corpus, the number of terms in the document, and
the sum of all term counts in the corpus.

We now turn to features based on {\it Cohesion}. Text cohesion
\cite{Halliday} involves relations between words or referring
expressions, which determine how tightly connected the text is.
Cohesion is brought about by linguistic devices such as repetition,
synonymy, anaphora, and ellipsis. Models of text cohesion have been
explored in application to information retrieval \cite{Salton}, where
paragraphs which are similar above some threshold to many other
paragraphs, i.e., ``bushy'' nodes, are considered likely to be
salient.  Text cohesion has also been applied to the explication of
discourse structure \cite{Morris}, \cite{Hearst}, and has been the
focus of renewed interest in text summarization
\cite{Boguraev}, \cite{Mani},\cite{Elhadad}. In our work, we use two cohesion-based features:
synonymy, and co-occurrence based on bigram statistics. To compute
synonyms the algorithm uses WordNet \cite{WordNet}, comparing
contentful nouns (their contentfulness determined by a
``function-word'' stoplist) as to whether they have a synset in common
(nouns are extracted by the Alembic part-of-speech tagger
\cite{Aberdeen}). Co-occurrence scores between contentful words up to 40 words apart are
computed using a standard mutual information metric \cite{Fano},
\cite{Church}: the mutual information between terms j and k in document i is:

{\small
\[ mutinfo(j, k, i) = \ln(\frac{n_{i}tf_{ji, ki}}{tf_{ji}tf_{ki}})  \]}

where $tf_{ji, ki}$ = maximum frequency of bigram $jk$ in document
$i$, $tf_{ji}$ = frequency of term $j$ in document i, $n_{i}$ = total
number of terms in document $i$.  The document in question is the
entire cmp-lg corpus.  The co-occurrence table only stores scores for
$tf$ counts greater than 10 and mutinfo scores greater than 10.

\section {Training Data}

We use a training corpus of computational linguistics texts. These are
198 full-text articles and (for the generic summarizer) their
author-supplied abstracts, all from the Computation and Language
E-Print Archive \cite{cmp-lg}, provided in SGML form by the University
of Edinburgh.  The articles are between 4 and 10 pages in length and
have figures, captions, references, and cross-references replaced by
place-holders. The average compression rate for abstracts in this corpus is
5\%.

\begin
{table*}
\centering
{\small
\begin{tabular}{|l|l|}\hline
{\bf Metric \/} &   {\bf Definition \/}  \\ \hline
Predictive Accuracy & Number of testing examples classified correctly \\ &  $/$ total
number of  test examples.  \\ \hline
Precision & Number of positive examples classified correctly \\ &  $/$ number of examples
classified positive, during testing \\ \hline
Recall & Number of positive examples classified correctly \\ &  $/$ Number known positive,
during testing \\ \hline
(Balanced) F-score & $2(Precision \cdot Recall) / (Precision + Recall)$ \\ \hline
\end{tabular}}
\caption{Metrics used to measure learning performance}
\label{table-3}
\end{table*}

Once the sentences in each text (extracted using a sentence tagger
\cite{Aberdeen}) are coded as feature vectors, they are labeled with
respect to relevance to the text's abstract. The labeling function
uses the following similarity metric:

{\small
\[ N1 + \frac{\sum_{i=1}^{N2} i_{s1}.i_{s2}}{\sqrt{\sum_{i=1}^{N2} i^{2}_{s1} . \sum_{i=1}^{N2} i^{2}_{s2}}} \]
} where $i_{s1}$ is the tf.idf weight of word i in sentence s1, N1 is the
number of words in common between s1 and s2, and N2 is the total
number of words in s1 and s2.

In labeling, the top $c$\% (where $c$ is the compression rate) of the
relevance-ranked sentences for a document are then picked as positive
examples for that document. This resulted in 27,803 training vectors,
with considerably redundancy among them, which when removed yielded
900 unique vectors (since the learning implementations we used ignore
duplicate vectors), of which 182 were positive and the others
negative.  The 182 positive vectors along with a random subset of 214
negative were collected together to form balanced training data of 396
examples. The labeled vectors are then input to the learning
methods.

Some preliminary data analysis on the ``generic'' training data
indicates that except for the two cohesion features, there is a
significant difference between the summary and non-summary counts for
some feature value of each feature ($\chi^{2}$ test, $\rho$ $<$
0.001). This suggests that this is a reasonable set of features for
the problem, even though different learning methods may disagree on
importance of individual features.

\subsection {Generating user-focused training abstracts}

\label{user}

The overall information need for a user is defined by a set of
documents. Here a subject was told to pick a sample of 10 documents
from the cmp-lg corpus which matched his interests.  The top content
words were extracted from each document, scored by the $G^{2}$ score
(with the cmp-lg corpus as the background corpus). Then, a centroid
vector for the 10 user-interest documents was generated as
follows. Words for all the 10 documents were sorted by their scores
(scores were averaged for words occurring in multiple documents). All
words more than 2.5 standard deviations above the mean of these words'
scores were treated as a representation of the user's interest, or
topic (there were 72 such words). Next, the topic was used in a
spreading activation algorithm based on \cite{Mani} to discover, in
each document in the cmp-lg corpus, words related to the topic.

\begin
{table*}
\centering
{\small
\begin{tabular}{|l|l|l|l|l|}\hline
{\bf Method \/} & {\bf Predictive Accuracy \/} &   {\bf Precision \/} & {\bf Recall \/} & {\bf F-score \/}  \\ \hline
SCDF (Generic) & .64  & .66  & .58  & .62 \\ \hline
SCDF (User-Focused) & .88 & .88 & .89 & .88 \\ \hline
AQ (Generic) & .56 & .49 & .56 &  .52 \\ \hline
AQ (User-Focused) & .81 & .78 & .88 & .82  \\ \hline
C4.5 Rules (Generic, pruned) & .69 & .71 & .67 & .69 \\ \hline
C4.5 Rules (User-Focused, pruned) & .89  & .88 & .91 & .89   \\ \hline 
\end{tabular}}
\caption{Accuracy of learning algorithms (at 20\% compression)}
\label{table-4}
\end{table*}

Once the words in each of the corpus documents have been reweighted by
spreading activation, each sentence is weighted based on the average
of its word weights. The top $c$\% (where $c$ is the compression rate)
of these sentences are then picked as positive examples for each
document, together constituting a user-focused abstract (or extract)
for that document. Further, to allow for user-focused features to be
learnt, each sentence's vector is extended with two additional
user-interest-specific features: the number of reweighted words
(called {\it keywords}) in the sentence as well as the number of
keywords per contentful word in the sentence\footnote{We don't use
specific keywords as features, as we would prefer to learn rules which
could transfer across interests.}. Note that the {\it keywords}, while
including terms in the user-focused abstract, include many other
related terms as well.

\section {Learning Methods}

We use three different learning algorithms: Standardized Canonical Discriminant Function (SCDF) analysis
\cite{SPSS}, C4.5-Rules \cite{Quinlan}, and AQ15c \cite{Wnek}. 
SCDF is a multiple regression technique which creates a linear
function that maximally discriminates between summary and non-summary
examples. While this method, unlike the other two, doesn't have the
advantage of generating logical rules that can easily be edited by a
user, it offers a relatively simple method of telling us to what
extent the data is linearly separable.  

\subsection {Results}

The metrics for the learning algorithms used are shown in
Table~\ref{table-3}. In Table~\ref{table-4}, we show results averaged
over ten runs of 90\% training and 10\% test, where the test data
across runs is disjoint\footnote{SCDF uses a holdout of 1
document.}.

Interestingly, in the C4.5 learning of generic summaries on this
corpus, the thematic and cohesion features are referenced mostly in
rules for the negative class, while the location and tf features are
referenced mostly in rules for the positive class. In the user-focused
summary learning, the number of {\it keywords} in the sentence is the
single feature responsible for the dramatic improvement in learning
performance compared to generic summary learning; here the rules have
this feature alone or in combination with tests on locational
features.  User-focused interests tend to use a subset of the
locational features found in generic interests, along with
user-specific keyword features.

Now, SCDF does almost as well as C4.5 Rules for the user-focused case. This is
because the {\it keywords} feature is most influential in rules learnt by
either algorithm. However, not all the positive user-focused examples which
have significant values for the {\it keywords} feature are linearly separable
from the negative ones; in cases where they aren't, the other algorithms yield
useful rules which include {\it keywords} along with other features.  In the
generic case, the positive examples are linearly separable to a much lesser
extent. 

Overall, although our figures are higher the 42\% reported by Parc,
their performance metric is based on overlap between positively
labeled sentences and individual sentences in the abstract, whereas
ours is based on overlap with the abstract as a whole, making it
difficult to compare.  It is worth noting that the most effective
features in our generic learning are a subset of the Parc features,
with the cohesion features contributing little to overall
performance. However, note that unlike the Parc work, we do not avail
of ``indicator'' phrases, which are known to be genre-dependent.  In
recent work using a similar overlap metric, \cite{Teufel} reports that
the indicator phrase feature is the single most important feature for
accurate learning performance in a sentence extraction task using this
corpus; it is striking that we get good learning performance without
exploiting this feature.

 
Analysis of C4.5-Rules learning curves generated at 20\% compression reveal
some learning improvement in the generic case - (.65-.69) Predictive Accuracy,
and (.64-.69) F-Score, whereas the user-focused case reaches a plateau very
early - (.88-.89) Predictive Accuracy and F-Score. This again may be attributed
to the relative dominance of the {\it keyword} feature. We also found
surprisingly little change in learning performance as we move from 5\% to 30\%
compression.  These latter results suggests that this approach maintains high
performance over a certain spectrum of summary sizes. Inspection of the rules
shows that the learning system is learning similar rather than different rules
across compression rates. 

Some example rules are as follows:

{\small
\begin{verbatim} If the sentence is in the conclusion 
and it is a high tf.idf sentence, 
then it is a summary sentence.
(C4.5 Generic Rule 20, run 1, 20% compression.) 

If the sentence is in a section of depth 2
and the number of keywords is between 5 and 7
and the keyword to content-word ratio is 
between 0.43 and 0.58 (inclusive),
then it is a summary sentence. 
(AQ User-Focused Rule 7, run 4, 5% compression.)
\end{verbatim}}

As can be seen, the learnt rules are highly intelligible, and thus are
easily edited by humans if desired, in contrast with approaches (such
as SCDF or naive Bayes) which learn a mathematical function. In practice, this
becomes useful because a human may use the learning methods to generate an
initial set of rules, whose performance may then be evaluated on the data
as well as against intuition, leading to improved performance.

\section{Conclusion}

We have described a corpus-based machine learning approach to produce
generic and user-specific summaries. This approach shows encouraging
learning performance. The rules learnt for user-focused interests tend
to use a subset of the locational features found in rules for generic
interests, along with user-specific keyword features. The rules are
intelligible, making them suitable for human use. The approach is
widely applicable, as it does not require manual tagging or
sentence-level text alignment. In the future, we expect to also
investigate the use of regression techniques based on a continuous
rather than boolean labeling function. Of course, since learning the
labeling function doesn't tell us anything about how useful the
summaries themselves are, we plan to carry out a (task-based)
evaluation of the summaries. Finally, we intend to apply this approach
to other genres of text, as well as languages such as Thai and
Chinese.

\section* {Acknowledgments}
We are indebted to Simone Teufel, Marc Moens and Byron Georgantopoulos
(University of Edinburgh) for providing us with the cmp-lg corpus, and
to Barbara Gates (MITRE) for helping with the co-occurrence data.

\end{document}